\documentstyle[proceedings,epsfig,amstex,amssymb, floatflt]{crckapb}
\newtheorem{Pro}{Property}
\newtheorem{Lem}{Lemma}

\newtheorem{Def}{Definition}

\newcommand{\be}{\begin{equation}}
\newcommand{\ee}{\end{equation}}
\newcommand{\beq}{\begin{equation}}
\newcommand{\eeq}{\end{equation}}
\newcommand{\bed}{\begin{displaymath}}
\newcommand{\eed}{\end{displaymath}}
\newcommand{\beqa}{\begin{eqnarray}}
\newcommand{\eeqa}{\end{eqnarray}}
\newcommand{\beqann}{\begin{eqnarray*}}
\newcommand{\eeqann}{\end{eqnarray*}}
\newcommand{\bseq}{\begin{subequations}}
\newcommand{\eseq}{\end{subequations}}

\newcommand{\ba}{\begin{array}}
\newcommand{\ea}{\end{array}}

\newcommand{\set}[3]{\{\,#1\,\}_{#2}^{#3}}

\newcommand{\tr}{{\rm tr}}

\newcommand{\negr}[1]{{\bf {#1}}}
\newcommand{\nigr}[2]{{\bf {#1}}_{#2}}

\begin{opening}
\title{ON ISOTROPIC SETS OF POINTS IN THE PLANE.
\goodbreak\noindent\bf APPLICATION TO THE DESIGN OF ROBOT ARCHITECTURES}
\author{J. ANGELES}
\institute{Department of Mechanical Engineering, Centre for
Intelligent Machines, Montreal, Canada \\
 {\tt email: angeles\symbol{64}cim.mcgill.ca}}
\author{D. CHABLAT}
\institute{Institut de Recherche en Communication et
Cybern\'etique de Nantes, Nantes, France}
\end{opening}

\begin{document}
\maketitle
\parindent=0pt
\pagestyle{empty} {\bf Abstract:} Various performance indices are
used for the design of serial manipulators. One method of
optimization relies on the condition number of the Jacobian
matrix. The minimization of the condition number leads, under
certain conditions, to isotropic configurations, for which the
roundoff-error amplification  is lowest. In this paper, the
isotropy conditions, introduced elsewhere, are the motivation
behind the introduction of isotropic sets of points. By connecting
together these points, we define families of isotropic
manipulators. This paper is devoted to planar manipulators, the
concepts being currently extended to their spatial counterparts.
Furthermore, only manipulators with revolute joints are considered
here.

%%%%%%%%%%%%%%%%%%%%%%%%%%%%%%%%%%%%%%%%%%%%%%%%
\section{Introduction}
%%%%%%%%%%%%%%%%%%%%%%%%%%%%%%%%%%%%%%%%%%%%%%%%
Various performance indices have been devised to assess the
kinetostatic performance of serial manipulators. The literature on
performance indices is extremely rich to fit in the limits of this
paper, the interested reader being invited to look at it in the
rather recent references cited here. A dimensionless quality index
was recently introduced by Lee, Duffy, and Hunt (1998) based on
the ratio of the Jacobian determinant to its maximum absolute
value, as applicable to parallel manipulators. This index does not
take into account the location of the operation point in the
end-effector, for the Jacobian determinant is independent of this
location. The proof of the foregoing fact is available in
(Angeles, 1997), as pertaining to serial manipulators, its
extension to their parallel counterparts being straightforward.
The {\em condition number} of a given matrix, on the other hand is
well known to provide a measure of invertibility of the matrix
(Golub and Van Loan, 1989). It is thus natural that this concept
found its way in this context. Indeed, the condition number of the
Jacobian matrix was proposed by Salisbury and Craig (1982) as a
figure of merit to minimize when designing manipulators for
maximum accuracy. In fact, the condition number gives, for a
square matrix, a measure of the relative roundoff-error
amplification of the computed results (Golub and Van Loan, 1989)
with respect to the data roundoff error. As is well known,
however, the dimensional inhomogeneity of the entries of the
Jacobian matrix prevents the straightforward application of the
condition number as a measure of Jacobian invertibility. The {\em
characteristic length} was introduced in (Angeles and
L\'opez-Caj\'un, 1992) to cope with the above-mentioned
inhomogeneity. Apparently, nevertheless, this concept has found
strong opposition within some circles, mainly because of the lack
of a direct geometric interpretation of the concept. It is the aim
of this paper to shed more light in this debate, by resorting to
the concept of {\em isotropic sets of points}. Briefly stated, the
application of isotropic sets of points to the design of
manipulator architectures relies on the concept of {\em distance}
in the space of $m\times n$ matrices, which is based, in turn, on
the Frobenius norm of matrices. With the purpose of rendering the
Jacobian matrix dimensionally homogeneous, moreover, we introduce
the concept of posture-dependent {\em conditioning length}. Thus,
given an arbitrary serial manipulator in an arbitrary posture, it
is possible to define a unique length that renders this matrix
dimensionally homogeneous and of minimum distance to isotropy. The
characteristic length of the manipulator is then defined as the
conditioning length corresponding to the posture that renders the
above-mentioned distance a minimum over all possible manipulator
postures.

\hskip .7cm
It is noteworthy that isotropy comprising
symmetry at its core, manipulators with only revolute joints are
considered here. It should be apparent that mixing actuated
revolutes with actuated prismatic joints would destroy symmetry,
and hence, isotropy.

%%%%%%%%%%%%%%%%%%%%%%%%%%%%%%
\section{Algebraic Background}
%%%%%%%%%%%%%%%%%%%%%%%%%%%%%%
When comparing two {\em dimensionless} $m\times n$ matrices \negr
A and \negr B, we can define the distance $d(\negr A,\,\negr B)$
between them as the Frobenius norm of their difference, namely,
 \vskip -0.8cm
 \begin{equation}
   d(\negr A,\,\negr B)\equiv \|\negr A - \negr B\|
   \label{e:distance}
  {\rm ~~i.e.,~~}
   d(\negr A,\,\negr B)\equiv
   \sqrt{\frac{1}{n}{\rm tr}
   [(\negr A - \negr B) (\negr A - \negr B)^T]}
   \label{e:distance-explicit}
  \end{equation}
An $m\times n$ isotropic matrix, with $m<n$, is one with a
singular value $\sigma>0$ of multiplicity $m$, and hence, if the
$m\times n$ matrix \negr C is isotropic, then
 \begin{equation}
  \negr C\negr C^T = \sigma^2\negr 1
  \label{e:isotropic}
 \end{equation}
where \negr 1 is the $m\times m$ identity matrix. Note that the
generalized inverse of \negr C can be computed without
roundoff-error, for it is proportional to $\negr C^T$, namely,
 \vskip -0.9cm
 \begin{equation}
  (\negr C\negr C^T)^{-1}\negr C^T=\frac{1}{\sigma^2}\negr C^T
  \label{e:gen-inverse}
 \end{equation}
\begin{figure}
\begin{floatingfigure}[r]{41mm}
\begin{center}
 \includegraphics[width=41mm,height=32mm]{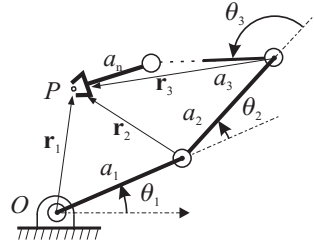}
 \vskip -0.3cm
  \caption{Planar $n$-revolute manipulator}
 \protect\label{figure:planar_n_R_manipulator}
\end{center}
\end{floatingfigure}
\end{figure}
Furthermore, the condition number $\kappa(\negr A)$ of a square
matrix \negr A is defined as (Golub and Van Loan, 1989)
 \begin{eqnarray}
  \kappa(\negr A)= \|\negr A\| \|\negr A^{-1}\|
  \label{equation:kappa}
 \end{eqnarray}
where {\em any} norm can be used. For purposes of the paper, we
shall use the Frobenius norm for matrices and the Euclidean norm
for vectors. Henceforth we assume, moreover, a planar $n$-revolute
manipulator, as depicted in
Fig.~\ref{figure:planar_n_R_manipulator}, with Jacobian matrix
\negr J given by (Angeles, 1997)
 \begin{eqnarray}
  \negr J=
  \left[\begin{array}{cccc} 1 & 1 & \cdots & 1 \\ \negr E \negr r_1
  & \negr E \negr r_2 & \cdots & \negr E \negr r_n
       \end{array}
   \right] \label{equation:jacobian_matrix} {\rm ~~,~~~~~} \negr E=
  \left[\begin{array}{cc} 0 & -1 \\ 1 & 0
  \end{array}
  \right] \label{e:E}
 \end{eqnarray}
where $\negr r_i$ is the vector directed from the center of the
$i$th revolute to the operation point $P$ of the end-effector, and
matrix $\negr E$  represents a counterclockwise rotation of
$90^\circ$.

\hskip .7cm It will prove convenient to partition {\negr J} into a
$1 \times n$ block \negr A and a $2 \times n$ block \negr B,
defined as $\negr A = [1~1~\cdots~1]$ and $\negr B = [\negr E
\negr r_1~\negr E \negr r_2~ \cdots~\negr E \negr r_n]$.
Therefore, while the entries of \negr A are dimensionless, those
of \negr B have units of length. Thus, the sole singular value of
\negr A, i.e., the nonnegative square root of the scalar of $\negr
A \negr A^T$, is $\sqrt{n}$, and hence, dimensionless, and
pertains to the mapping from joint-rates into end-effector angular
velocity. The singular values of \negr B, which are the
nonnegative square roots of the eigenvalues of $\negr B \negr
B^T$, have units of length, and account for the mapping from
joint-rates into operation-point velocity. It is thus apparent
that the singular values of $\negr J$ have different dimensions
and hence, it is impossible to compute $\kappa(\negr J)$ as in
eq.(\ref{equation:kappa}), for the norm of $\negr J$ cannot be
defined. The normalization of the Jacobian for purposes of
rendering it dimensionless has been judged to be dependent on the
normalizing length (Paden, and Sastry, 1988; Li, 1990). As a means
to avoid the arbitrariness of the choice of that normalizing
length, the characteristic length $L$ was introduced in
(Ranjbaran, Angeles, Gonz\'alez-Palacios, and Patel, 1995). We
shall resort to this concept, while shedding more light on it, in
discussing manipulator architectures.

%%%%%%%%%%%%%%%%%%%%%%%%%%%%%%
\section{Isotropic Sets of Points}
%%%%%%%%%%%%%%%%%%%%%%%%%%%%%%
Consider the set ${\cal S}\equiv \set{P_k}1n$ of $n$ points in the
plane, of position vectors $\set{\negr p_k}1n$, and centroid $C$,
of position vector \negr c, i.e.,
 \begin{equation}
  \negr c\equiv\frac{1}{n}\sum_1^n \negr p_k
  \label{e:centroid}
 \end{equation}
\hskip .7cm The summation appearing in the right-hand side of the
above expression is known as the {\em first moment of $\cal S$
with respect to the origin $O$} from which the position vectors
stem. The {\em second moment of $\cal S$ with respect to $C$} is
defined as a tensor \negr M, namely,
 \begin{equation}
 \negr M\equiv
 \sum_1^n (\nigr pk-\negr c)(\nigr pk-\negr c)^T
 \label{e:second}
 \end{equation}
\hskip .7cm It is now apparent that the {\em root-mean square}
value of the distances $\set{d_k}1n$ of $\cal S$, $d_{\rm rms}$,
to the centroid is directly related to the trace of \negr M,
namely,
 \begin{equation}
  d_{\rm rms}\equiv
  \sqrt{\frac{1}{n}\sum_1^n (\nigr pk-\negr c)^T
  (\nigr pk-\negr c)} \equiv \sqrt{\frac{1}{n}\tr(\negr M)}
  \label{e:d_rms}
 \end{equation}
\hskip .7cm Further, the {\em moment of inertia} \negr I of $\cal
S$ with respect to the centroid is defined as that of a set of
unit masses located at the points of $\cal S$, i.e.,
 \begin{subequations}
  \begin{equation}
   \negr I\equiv
   \sum_1^n [\,\|\nigr pk-\negr c\|^2\negr 1 -
   (\nigr pk -\negr c)(\nigr pk-\negr c)^T\,]
   \label{e:inertia}
 \end{equation}
in which \negr 1 is the $3\times 3$ identity matrix. Hence, in
light of definitions (\ref{e:second}) and (\ref{e:d_rms}),
 \vskip -1.0cm
 \begin{equation}
  \negr I= \tr(\negr M) \negr 1 - \negr M
  \label{e:inertia-alt}
 \end{equation}
 \end{subequations}
\hskip .7cm We shall refer to {\bf I} as the {\em geometric moment
of inertia} of $\cal S$ about its centroid. It is now apparent
that \negr I is composed of two parts, an isotropic matrix of norm
$\tr(\negr M)$ and the second moment of $\cal S$ with the sign
reversed. Moreover, the moment of inertia \negr I can be expressed
in a form that is more explicitly dependent upon the set
$\set{\nigr pk- \negr c}1n$, if we recall the concept of {\em
cross-product matrix} (Angeles, 1997): For any three-dimensional
vector \negr v, we define the cross-product matrix $\nigr Pk$ of
$(\nigr pk - \negr c)$, or of any other three-dimensional vector
for that matter, as
 \begin{subequations}
  \begin{equation}
   \nigr Pk\equiv
   \frac{\partial [(\nigr pk-\negr c)\times\negr v]}
        {\partial\negr v}
   \label{e:CPM}
  \end{equation}
\hskip .7cm Further, we recall the identity (Angeles, 1997)
 \begin{equation}
  \negr P_k^2\equiv
  -\|\nigr pk-\negr c\|^2\negr 1 +
  (\nigr pk-\negr c) (\nigr pk-\negr c)^T
  \label{e:CPM^2}
  \end{equation}
 \end{subequations}
\hskip .7cm It is now apparent that the moment of inertia of $\cal
S$ takes the simple form
 \vskip -1.0cm
 \begin{equation}
  \negr I = - \sum_1^n \negr P_k^2
  \label{e:inertia-CPM}
 \end{equation}
\vskip -0.3cm \hskip .7cm We thus have \begin{Def}[Isotropic Set]
The set $\cal S$ is said to be isotropic if its second-moment
tensor with respect to its centroid is isotropic.
\end{Def}
 \hskip .7cm
As a consequence, we have
\begin{Lem}
The geometric moment of inertia of an isotropic set of points is
isotropic. Conversely, an isotropic geometric moment of inertia
pertains necessarily to an isotropic set of points.
\end{Lem}
We describe below some properties of isotropic sets of points that
will help us better visualize the results that follow.

%%%%%%%%%%%%%%%%%%%%%%%%%%%%%%
\subsection{ISOTROPY-PRESERVING OPERATIONS ON SETS OF POINTS}
%%%%%%%%%%%%%%%%%%%%%%%%%%%%%%
Consider two isotropic sets of points in the plane, ${\cal S}_1=
\set{P_k}1n$ and ${\cal S}_2=\set{P_k}{n+1}{n+m}$. If the centroid
$C$ of the position vector $\negr c$ of ${\cal S}_1$ coincides
with that of ${\cal S}_2$, i.e. if,
 \begin{eqnarray}
  \negr c \equiv
  \frac{1}{n} \sum_1^n \negr p_k \equiv
  \frac{1}{m}\sum_{n+1}^{n+m} \negr p_k
 \end{eqnarray}
then, the set ${\cal S}={\cal S}_1 \cup {\cal S}_2$ is isotropic.
Hence,
\begin{Pro}
The union of two isotropic sets of points sharing the same
centroid is also isotropic.
\end{Pro}
\hskip .7cm Furthermore, as the reader can visualize, we state
below one more operation on sets of points, namely, a rigid-body
rotation, that preserves isotropy:
\begin{Pro}
The rotation of an isotropic set of points as a rigid body with
respect to its centroid is also isotropic.
\end{Pro}

%%%%%%%%%%%%%%%%%%%%%%%%%%%%%%
\subsection{TRIVIAL ISOTROPIC SETS OF POINTS}
%%%%%%%%%%%%%%%%%%%%%%%%%%%%%%
An isotropic set of points can be defined by the union, rotation,
 or a combination of both, of isotropic sets. The simplest
set of isotropic points is the set of vertices of a regular
polygon. We thus have
\begin{Def}[Trivial isotropic set]
A set of $n$ points ${\cal S}$ is called {\em trivial} if it is
the set of vertices of a regular polygon with $n$ vertices.
\end{Def}
\begin{Lem}
A trivial isotropic set ${\cal S}$ remains isotropic under every
reflection about an axis passing through the centroid $C$.
\end{Lem}

%%%%%%%%%%%%%%%%%%%%%%%%%%%%%%
\section{An Outline of Kinematic Chains}
%%%%%%%%%%%%%%%%%%%%%%%%%%%%%%
The connection between sets of points and planar manipulators of
the serial type is the concept of {\em simple kinematic chain}.
For completeness, we recall here some basic definitions pertaining
to this concept.

%%%%%%%%%%%%%%%%%%%%%%%%%%%%%%%%%%%%%%%%
\subsection{SIMPLE KINEMATIC CHAINS}
%%%%%%%%%%%%%%%%%%%%%%%%%%%%%%%%%%%%%%%%
The kinematics of manipulators is based on the concept of {\em
kinematic chain}. A kinematic chain is a set of {\em rigid
bodies}, also called {\em links}, coupled by {\em kinematic
pairs}. In the case of planar chains, two lower kinematic pairs
are possible, the revolute, allowing pure rotation of the two
coupled links, and the prismatic pair, allowing a pure relative
translation, along one direction, of the same links. For the
purposes of this paper, we study only revolute pairs, but
prismatic pairs are also common in manipulators.
\par
\begin{Def}[Simple kinematic chain]
A kinematic chain is said to be {\em simple} if each and every one
of its links is coupled to at most two other links.
\end{Def}
\hskip .7cm A simple kinematic chain can be {\em open} or {\em
closed}; in studying serial manipulators we are interested in the
former. In such a chain, we distinguish exactly two links, the
terminal ones, coupled to only one other link. These links are
thus said to be {\em simple}, all other links being {\em binary}.
In the context of manipulator kinematics, one terminal link is
arbitrarily designated as {\em fixed}, the other terminal link
being the {\em end-effector} (EE), which is the one executing the
task at hand. The task is defined, in turn, as a sequence of {\em
poses}---positions and orientations---of the EE, the position
being given at a specific point $P$ of the EE that we term the
{\em operation point}.

%%%%%%%%%%%%%%%%%%%%%%%%%%%%%%%%%%%%%%%%
\subsection{ISOTROPIC KINEMATIC CHAINS}
%%%%%%%%%%%%%%%%%%%%%%%%%%%%%%%%%%%%%%%%
To every set $\cal S$ of $n$ points it is possible to associate a
number of kinematic chains. To do this, we number the points from
1 to $n$, thereby defining $n-1$ links, the $i$th link carrying
joints $i$ and $i+1$. Links are thus correspondingly numbered from
1 to $n$, the $n$th link, or EE, carrying joint $n$ on its
proximal (to the base) end and the operation point $P$ on its
distal end. Furthermore, we define an additional link, the base,
which is numbered as 0.
\par
\hskip .7cm It is now apparent that, since we can number a given
set $\cal S$ of $n$ points in $n!$ possible ways, we can associate
$n!$ kinematic chains to the above set $\cal S$ of $n$ points.
Clearly, these chains are, in general, different, for the lengths
of their links are different as well. Nevertheless, some pairs of
identical chains in the foregoing set are possible.
\begin{Def}[Isotropic kinematic chain]
Let set $\cal S$ of $n$ points be iso-tropic, and the operation
point $P$ be defined as the centroid of $\cal S$. Any kinematic
chain stemming from $\cal S$ is said to be isotropic.
\end{Def}
%%%%%%%%%%%%%%%%%%%%%%%%%%%%%%%%%%%%%%%%
\section{The Posture-Dependent Conditioning Length of
Planar n-Re-volute Manipulators}
%%%%%%%%%%%%%%%%%%%%%%%%%%%%%%%%%%%%%%%%
Under the assumption that the manipulator finds itself at a {\em
posture} $\cal P$ that is given by its set of joint angles,
$\set{\theta_k}1 n$, we start by dividing the last two rows of the
Jacobian by a length $l_{\cal P}$, as yet to be determined. This
length will be found so as to minimize the distance of the
normalized Jacobian to a corresponding isotropic matrix \negr K,
subscript $ \cal P$ reminding us that, as the manipulator changes
its posture, so does the length $l_{\cal P}$. This length will be
termed the {\em conditioning length} of the manipulator at $\cal
P$.

%%%%%%%%%%%%%%%%%%%%%%%%%%%%%%
\subsection{A DIMENSIONALLY-HOMOGENEOUS JACOBIAN MATRIX}
%%%%%%%%%%%%%%%%%%%%%%%%%%%%%%
In order to distinguish the original Jacobian matrix from its
dimensionally-homogeneous counterpart, we shall denote the latter
by $\overline{\negr J}$, i.e.,
 \begin{eqnarray}
  \overline{\negr J}=
  \left[\begin{array}{cccc}
   1 & 1 & \cdots & 1 \\
   (1/ l_{\cal P})~\negr E \negr r_1 &
   (1/ l_{\cal P})~ \negr E \negr r_2 &
   \cdots & (1/ l_{\cal P})~ \negr E \negr r_n
   \end{array}
   \right] \nonumber
 \end{eqnarray}
\hskip .7cm Now the conditioning length will be defined via the
minimization of the distance of the dimensionally-homogeneous
Jacobian matrix $\overline{\negr J}$ of an $n$-revolute
manipulator to an isotropic $3\times n$ {\em model matrix} \negr K
whose entries are dimensionless and has the same gestalt as any
$3\times n$ Jacobian matrix. To this end, we define an isotropic
set ${\cal K}=\set{K_i}1n$ of $n$ points in a {\em dimensionless}
plane, of position vectors $\set{\nigr ki}1n$, which thus yields
the dimensionless matrix
 \begin{eqnarray}
  \negr K= \left[\begin{array}{cccc}
   1 & 1 & \cdots & 1 \\
   \negr E\nigr k1 &
   \negr E\nigr k2 &
   \cdots &
   \negr E\negr k_n
  \end{array}
  \right]
  \label{e:K}
 \end{eqnarray}
\hskip .7cm Further, we compute the product $\negr K\negr K^T$:
 \begin{eqnarray}
  \negr K \negr K^T=
  \left[\begin{array}{cc}
   n &
   \sum_1^n \negr k_i^T\negr E^T \\
   \sum_1^n \negr E\negr k_i &
   \sum_1^n \negr E\negr k_i\negr k_i^T\negr E^T
  \end{array}
  \right] \nonumber
 \end{eqnarray}
\hskip .7cm Upon expansion of the summations occurring in the
above matrix, we have
 \begin{eqnarray}
  \sum_1^n \negr k_i^T\negr E^T =
  (\sum_1^n \negr E \negr k_i)^T =
  \negr E(\sum_1^n \nigr ki)^T {\rm ~and~}
   \sum_1^n \negr E\negr k_i\negr k_i^T\negr E^T=
   \negr E(\sum_1^n \negr k_i\negr k_i^T)\negr E^T
  \nonumber
 \end{eqnarray}
\hskip .7cm Now, by virtue of the assumed isotropy of $\cal K$,
the terms in parentheses in the foregoing expressions become
 \begin{eqnarray}
  \sum_1^n \nigr ki=\negr 0 {\rm ~~~and ~~~~} %%\nonumber\\
  \sum_1^n \nigr ki\negr k_i^T=k^2\nigr 1{2\times 2}\nonumber
 \end{eqnarray}
where the factor $k^2$ is as yet to be determined and $\nigr
1{2\times 2}$ denotes the $2\times 2$ identity matrix. Hence, the
product $\negr K\negr K^T$ takes the form
 \vskip -0.3cm
 \begin{eqnarray}
  \negr K\negr K^T=
  \left[\begin{array}{cc}
           n & \negr 0^T \\
           \negr 0 & k^2\nigr 1{2\times 2}
        \end{array}
  \right]
  \label{e:KK^T}
 \end{eqnarray}
 \vskip -0.2cm
\hskip .7cm Now, in order to determine $k^2$, we recall that
matrix \negr K is isotropic, and hence that the product $\negr
K\negr K^T$ has a triple eigenvalue. It is now apparent that the
triple eigenvalue of the said product must be $n$, which means
that
 \vskip -0.5cm
 \begin{eqnarray}
  k^2=n\label{e:k^2} {\rm ~~~and~~~}
  \sum_1^n \nigr ki\negr k_i^T=(n)\nigr 1{2\times 2}
 \end{eqnarray}
 \vskip -0.5cm

%%%%%%%%%%%%%%%%%%%%%%%%%%%%%%
\subsection{COMPUTATION OF THE CONDITIONING LENGTH}
%%%%%%%%%%%%%%%%%%%%%%%%%%%%%%
We can now formulate a least-square problem aimed at finding the
conditioning length $l_{\cal P}$ that renders the distance from
$\overline{\negr J}$ to \negr K a minimum. The task will be eased
if we work rather with the reciprocal of $l_{\cal P}$,
$\lambda\equiv 1/l_{\cal P}$, and hence,
 \vskip -0.6cm
 \begin{equation}
  z\equiv \frac{1}{2}
  \frac{1}{n}
  \tr[(\overline{\negr J}-\negr K)
  (\overline{\negr J}-\negr K)^T] \quad\rightarrow\quad\min_\lambda
  \label{e:least-squares}
 \end{equation}
\hskip .7cm Upon simplification, and recalling that $\tr(\negr
K\overline{\negr J}^T)= \tr(\overline{\negr J} \negr K^T)$,
 \begin{equation}
  z\equiv \frac{1}{2}
  \frac{1}{n}
  \tr(\overline{\negr J}\overline{\negr J}^T
  -2\negr K\overline{\negr J}^T+\negr K\negr K^T)
  \label{e:simple-z}
 \end{equation}
\hskip .7cm It is noteworthy that the above minimization problem
is (a) quadratic in $\lambda$, for $\overline{\negr J}$ is linear
in $\lambda$ and (b) unconstrained, which means that the problem
accepts a unique solution. This solution can be found,
additionally, in closed form. Indeed, the optimum value of
$\lambda$ is readily obtained upon setting up the normality
condition of the above problem, namely,
 \begin{equation}
  \frac{\partial z}
  {\partial\lambda}\equiv \frac{1}{2n}{\rm tr} \left(\frac{\partial
  (\overline{\negr J}~\overline{\negr J}^T)} {\partial\lambda}
  \right) - \frac{1}{n}{\rm tr}\left(\negr K
  \frac{\partial\overline{\negr J}^T}{\partial\lambda}\right)=0
  \label{e:normal-1}
 \end{equation}
where we have used the linearity property of the trace and the
derivative operators. The normality condition then reduces to
 \begin{eqnarray}
  \lambda \sum_1^n\|\negr r_j\|^2 -
  \sum_1^n \negr k_j^T\negr r_j=0 \nonumber
 \end{eqnarray}
\hskip .7cm Now, if we notice that $\|\negr r_j\|$ is the distance
$d_j$ from the operation point $P$ to the center of the $j$th
revolute, the first summation of the above equation yields $n
d_{\rm rms}^2$, with $d_{\rm rms}$ denoting the root-mean-square
value of the set of distances $\set{d_j}1 n$, and hence,
 \begin{eqnarray}
  \lambda =
  \frac{\sum_1^n \negr k_i^T\negr r_j}
  {nd_{\rm rms}^2}\quad\Rightarrow\quad l_{\cal P}=
  \frac{nd_{\rm rms}^2}{\sum_1^n \negr k_j^T\negr r_j}
  \label{e:l_P}
 \end{eqnarray}
\begin{Def}[Characteristic length]
The conditioning length of a manipulator with Jacobian matrix
capable of attaining isotropic values, as pertaining to the
isotropic Jacobian, is defined as the {\em characteristic length}
of the manipulator.
\end{Def}
\hskip .7cm An example is included below from which the reader can
realize that a reordering of the columns of \negr K preserves the
set of manipulators resulting thereof, and hence, \negr K is not
unique. Lack of space prevents us from including an example of a
nonisotropic manipulator. The interested reader can find such an
example in (Chablat and Angeles, 1999.)

\subsection{EXAMPLE: A FOUR-DOF REDUNDANT PLANAR MANIPULATOR}
%%%%%%%%%%%%%%%%%%%%%%%%%%%%%%
An isotropic set $\cal K$ of four points, $\set{K_i}14$, is
defined in a nondimensional plane, with the position vectors
$\negr k_i$ given below:
 \begin{subequations}
  \begin{equation}
   \negr k_1 =
  \left[\begin{array}{c}
             1 \\
             1
        \end{array}
  \right]
  {~\rm ,~}
  \negr k_2 =
  \left[\begin{array}{r}
           - 1 \\
             1
       \end{array}
  \right]
  {~\rm ,~}
  \negr k_3 =
  \left[\begin{array}{c}
           - 1 \\
           1
        \end{array}
  \right]
  {~\rm ,~}
  \negr k_4 =
  \left[\begin{array}{r}
            1 \\
            1
        \end{array}
  \right]
 \end{equation}
which thus lead to
 \begin{equation}
  \negr K=
  \left[\begin{array}{rrrr}
   1 &  1 &  1 &  1\\
  -1 & -1 &  1 &  1 \\
   1 & -1 & -1 &  1
  \end{array}
  \right]
  {\rm ~~~\Rightarrow~~~~}
  \negr K \negr K^T=
  \left[\begin{array}{cccc}
         4 &  0 & 0 \\
         0 &  4 & 0 \\
         0 &  0 & 4
        \end{array}
  \right]=
  \sigma^2 \negr 1
 \end{equation}
 \end{subequations}
 \begin{figure}[hbt]
  \vskip -0.3cm
  \begin{center}
    \includegraphics[width=97mm,height=53mm]{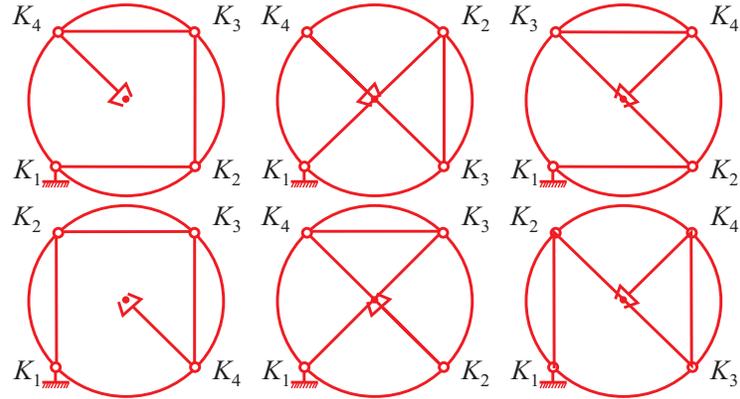}
    \caption{Six isotropic postures for the same isotropic set}
    \protect\label{figure:4dof_posture_isotrope}
  \end{center}
 \vskip -0.5cm
 \end{figure}
and is apparently isotropic, with a triple singular value of
$k=2$. We thus have $4! = 24$ isotropic kinematic chains for a
four-dof planar manipulator, but we represent only six in
Fig.~\ref{figure:4dof_posture_isotrope} because the choice of the
first point is immaterial, since this choice amounts to a rotation
of the overall manipulator as a rigid body. The conditioning
length $l_{\cal P}$ is equal to $l / 2$ for the three
manipulators, with $l$ denoting any nonzero length. Moreover, the
link lengths are defined, respectively, for three distinct
manipulators of Fig.~\ref{figure:4dof_posture_isotrope}, form left
to right: $a_1=l, a_2=l, a_3=l, a_4=\sqrt{2}l/2$; $a_1=l,
a_2=\sqrt{2}l, a_3=l, a_4=\sqrt{2}l/2$; $a_1=\sqrt{2}l, a_2=l,
a_3=\sqrt{2}l, a_4=\sqrt{2}l/2$.

\hskip .7cm An extensive discussion of isotropic sets of points
and the optimum design of manipulators is available in (Chablat
and Angeles, 1999.)

%%%%%%%%%%%%%%%%%%%%%%%%%%%%%%%%%%%%%%%%%%%%%%%%
\section{Conclusions}
%%%%%%%%%%%%%%%%%%%%%%%%%%%%%%%%%%%%%%%%%%%%%%%%
Isotropic manipulators are defined in this paper by resorting to
the concept of isotropic sets of points. This concept allows us to
define families of isotropic redundant manipulators. The
characteristic length can thus be defined as the conditioning
length $l_{\cal P}$ that pertains to the isotropic postures. The
paper focuses on planar manipulators, but the underlying concepts
are currently extended to their three-dimensional counterparts.

%%%%%%%%%%%%%%%%%%%%%%%%%%%%%%%%%%%%%%%%%%%%%%%%

\end{document}